\documentclass{amia}
\usepackage{graphicx}
\usepackage[labelfont=bf]{caption}
\usepackage[superscript,nomove]{cite}
\usepackage{color}
\usepackage{enumitem}
\usepackage[usestackEOL]{stackengine}
\usepackage{multicol, multirow}
\usepackage{wrapfig}
\usepackage[dvipsnames]{xcolor}
\usepackage[sort&compress,comma,numbers,super]{natbib}
\usepackage[rightcaption]{sidecap}
\usepackage{url}

\definecolor{forestgreen}{RGB}{34, 139, 34}

\begin{document}
\strutlongstacks{T}

\title{Leveraging Spatial Information in Radiology Reports for Ischemic Stroke Phenotyping}

\author{Surabhi Datta, MS$^1$, Shekhar Khanpara, MD$^2$, Roy F. Riascos, MD$^2$, Kirk Roberts, PhD$^1$}

\institutes{
    $^1$School of Biomedical Informatics,
    $^2$McGovern Medical School \\
    The University of Texas Health Science Center at Houston, Houston, TX
    % Houston, TX \\
}

\maketitle

\noindent{\bf Abstract}

\textit{
Classifying fine-grained ischemic stroke phenotypes relies on identifying important clinical information.
Radiology reports provide relevant information with context to determine such phenotype information.
We focus on stroke phenotypes with location-specific information: brain region affected, laterality, stroke stage, and lacunarity.
We use an existing fine-grained spatial information extraction system--Rad-SpatialNet--to identify clinically important information and apply simple domain rules on the extracted information to classify phenotypes.
The performance of our proposed approach is promising (recall of \textcolor{black}{$89.62$\%} for classifying brain region and \textcolor{black}{$74.11$\%} for classifying brain region, side, and stroke stage together).
Our work demonstrates that an information extraction system based on a fine-grained schema can be utilized to determine complex phenotypes with the inclusion of simple domain rules.
These phenotypes have the potential to facilitate stroke research focusing on post-stroke outcome and treatment planning based on the stroke location.
}

\section*{Introduction}
\label{sec:intro}

Ischemic stroke (IS) accounts for around $87$\% of all strokes in the United States \cite{2020StrokeFactsCdc}. 
Clinical trials and epidemiological studies \textcolor{black}{targeted toward investigating communication, cognitive, and emotional changes after stroke} are interested in analyzing specific subsets of patient records pertaining to certain characteristics of IS for treatment and prognosis research.
Radiological findings documented in head CT and brain MRI reports provide important information to develop IS phenotypes.
\textcolor{black}{Understanding and identifying various clinically important information from the report text can facilitate in constructing fine-grained phenotypes.
In this work, we propose to utilize spatial information in the reports to construct IS phenotypes.
We develop and evaluate a natural language processing (NLP) pipeline for IS phenotyping by using spatial information extracted from the reports.
More specifically, we use the spatially-related imaging features and their brain locations as well as the potential diagnoses information to classify the phenotypes.
}

\textcolor{black}{Effects of stroke in a patient are dependent on the areas of the brain affected \cite{EffectsStroke, hui2020IschemicStroke}.
Based on the side and the particular location of the stroke, different body functions are impaired.
For example, stroke in the right side of cerebral hemisphere results in \textit{left-sided weakness or paralysis}, \textit{visual}, and \textit{spatial problems}, stroke in the cerebellum manifests in a different set of effects such as \textit{ataxia}, \textit{dizziness}, \textit{nausea}, and \textit{vomiting}, whereas stroke located in the brainstem results in problems associated with \textit{breathing}, \textit{balance}, and \textit{coma}.
Moreover, the effects can be further specified based on the particular lobe of the cerebral hemisphere that is affected.
For example, \textit{sensation} and \textit{spatial awareness} are impacted with stroke in the parietal lobe whereas \textit{language} and \textit{memory} are impaired with stroke in the temporal lobe.
A previous work \cite{chengbastian2014InfluenceStrokeInfarct} has demonstrated that location of stroke infarct influences the functional outcome following an ischemic stroke as measured by modified Rankin Scale, a commonly used scale for rating stroke outcome in clinical trials.
Further, a few studies \cite{shi2017StudyBrainFunctional, price2010PredictingLanguageOutcome} have focused on the brain locations affected by stroke for improving treatment of post-stroke depression and predicting post-stroke language outcome.
}
\textcolor{black}{Therefore, categorizing imaging reports according to stroke location--or in other words, constructing phenotypes incorporating the stroke location--holds potential benefits for clinical research studies that focus on targeted treatment based on the specific brain region affected.}

\textcolor{black}{We construct the IS phenotypes by using the brain location information in the reports both directly and indirectly.
Direct use refers to including the side and the specific brain region affected by stroke in the phenotypes.
Indirect use of location includes deriving other crucial information such as stroke stage} 
\textcolor{black}{based on the particular brain region a certain imaging feature is detected.}
\textcolor{black}{Besides these, we also use the IS-related potential diagnoses information directly in the phenotypes}
\textcolor{black}{to extract the stroke stage in cases when it is included as part of the diagnosis phrase (e.g., \textit{subacute} stage in the diagnosis phrase `\textit{subacute infarction}').}

\begin{figure*}[t]
\includegraphics[width=0.7\textwidth]{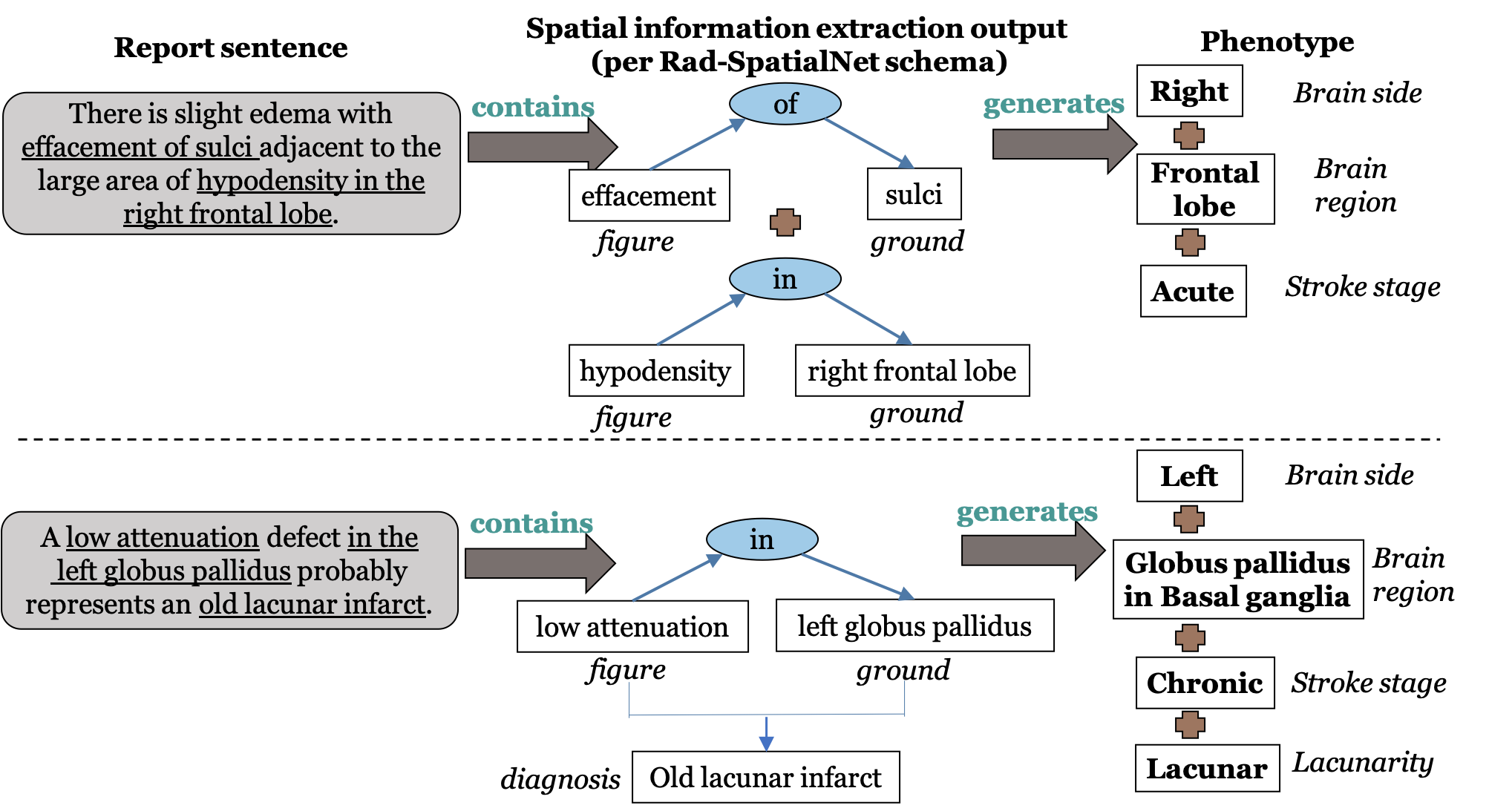}
\vspace{-0.1in}
\centering
\caption{Examples of stroke phenotypes using spatial relations from reports. Blue ovals contain spatial triggers.}
\label{fig:example_phenotypes}
\end{figure*}

\newpage
Consider the two examples shown in Figure \ref{fig:example_phenotypes} from head CT reports.
The first sentence captures information corresponding to mass effect like \textit{sulcal effacement} along with imaging feature such as \textit{cortical hypodensity} that helps to indicate that an \textit{infarction} is \textit{acute}. 
\textcolor{black}{The second sentence detects an area of \textit{low attenuation} in the left side of \textit{globus pallidus}, part of \textit{basal ganglia}.
The sentence also describes that this finding indicates that the infarct is \textit{lacunar} and thus \textit{chronic}.}
Therefore, for the first example, we see that spatial relations between imaging features and brain locations (as indicated by phrases like `\textit{effacement of sulci}' and `\textit{hypodensity in the right frontal lobe}') encode important radiological information that facilitates in determining the diagnosis (i.e., \textit{infarction}) and its stage (i.e., \textit{acute}).
Also, note that although \textit{acute} is not mentioned explicitly in this sentence, identifying the spatial relations help in inferring that the stroke is \textit{acute}.
Thus, spatial relations present in imaging reports can directly be utilized for constructing stroke phenotypes containing fine-grained location information along with additional derived information like stroke stage.
We, therefore, use our previously proposed spatial representation schema--Rad-SpatialNet \cite{datta2020RadSpatialNetFramebasedResource} to extract spatial information from reports which can subsequently be used for extracting important IS phenotypes.

\begin{SCfigure}
\includegraphics[width=0.75\textwidth]{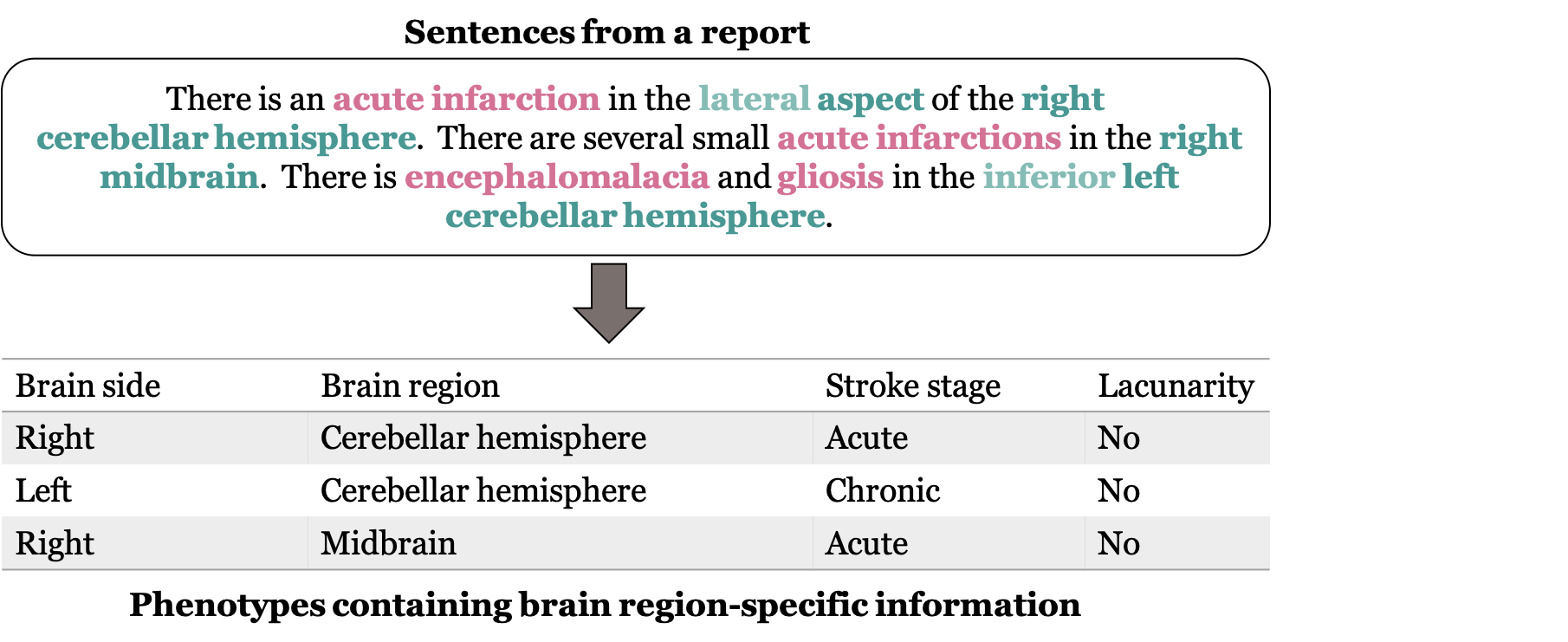}
% \vspace{-0.1in}
\centering
\caption{\textcolor{black}{Granular phenotypes considered in this work (shown for a sample report).}}
\label{fig:region_specific_info}
\end{SCfigure}

Prior studies have attempted to extract IS-related information from radiology reports. 
Wheater et al. \cite{wheater2019ValidatedNaturalLanguagea} developed brain imaging phenotypes, however, these phenotypes lacked specificity in the brain location information and were classified as only cortical or deep. 
Other works identified reports with acute IS \cite{ong2020MachineLearningNaturala, kim2019NaturalLanguageProcessinga} and silent brain infarcts \cite{fu2019NaturalLanguageProcessinga}.
However, these studies focused on limited information like classifying reports based on presence/absence of IS, acuity, and MCA territory involvement.
Alternatively, we aim to construct specific stroke phenotypes containing more granular information for each stroke affected brain area and this makes the task more complex compared to performing binary classification of the reports.
\textcolor{black}{We illustrate the granularity and complexity of our phenotypes in Figure \ref{fig:region_specific_info}. Note that the phenotypes consider information at the level of both side and region of the brain affected. Thus we see the stage is \textit{acute} for right cerebellar hemisphere and \textit{chronic} for the left side.}

\newpage
Therefore, using spatial information from the reports forms an intuitive way to extract such fine-grained information for constructing the phenotypes.
In this paper, we \textcolor{black}{define} the fine-grained stroke phenotypes described above with input from radiology experts.
For automatic labeling of the reports with the relevant phenotypes, we first identify the spatial relations using a transformer-based model (BERT \cite{devlin2019bert}) for each report.
We then apply rules based on domain knowledge on the extracted spatial information to classify the phenotypes.
Finally, we evaluate our system by comparing the automatically generated phenotypes with the gold phenotypes for a set of head CT and brain MRI reports.
\textcolor{black}{
Thus the main contributions of our work include:
\vspace{-0.15in}
\begin{itemize}
    \item Classify fine-grained ischemic stroke phenotypes by applying simple domain rules on top of spatial information extracted from neuroradiology reports.
    \vspace{-0.05in}
    \item Phenotypes contain information targeted at the level of a specific side and region of the brain affected.
\end{itemize}}

\section*{Related Work}
    Numerous works have focused on identifying certain subgroups of stroke patients using NLP techniques with the aim to facilitate timely patient triaging to select appropriate group of patients highly likely to encounter severe consequences. We highlight the relevant studies by categorizing them in the following three subsections: 
    
    \textbf{\textit{Identifying stroke/ischemic stroke}} Sedghi et al. \cite{sedghi2015MiningClinicalText} converted medical narratives to codified text based on expert provided sign and symptom phrases and they applied ML algorithms on the codified sentences to predict the presence of stroke in a patient.
    Majersik et al. \cite{majersikjenniferjAbstractWMP92HighPrecision} applied NLP-based approaches by adding context to n-grams that classified ischemic, hemorrhagic, and non-stroke cases with high precision by using different combination of clinical report types.
    A study by Kim et al. \cite{kim2019NaturalLanguageProcessinga} utilized document-feature matrix vectorization techniques to classify brain MRI reports for identifying acute ischemic stroke.
    Govindarajan et al. \cite{govindarajan2020ClassificationStrokeDisease} developed ML-based NLP approaches to identify whether the stroke is ischemic or hemorrhagic based on some pre-defined symptoms and patient factors.

    \textbf{\textit{Classifying stroke subtypes}} Two studies focused on automatically classifying stroke patients based on standard stroke subtype classification systems--the Trial of Org 10172 in Acute Stroke Treatment (TOAST) and the Oxfordshire Community Stroke Project (OCSP).
    Garg et al. \cite{garg2019AutomatingIschemicStroke} developed ML-based approaches to classify patients according to the TOAST ischemic stroke subtyping using neurology progress notes and neuroradiology reports for better patient management and outcome prediction.
    Sung et al. \cite{sung2020EMRbasedPhenotypingIschemic} constructed features based on the medical entities identified by MetaMap and then applied traditional ML techniques to classify stroke patients based on four clinical syndromes taken from OCSP classification system that considers the anatomical location of stroke.

    \textbf{\textit{Identifying stroke features}} 
    A recent study by Ong et al. \cite{ong2020MachineLearningNaturala} classified radiology reports based on three outcomes - presence of stroke, involvement of MCA location, and stroke acuity by using text featurization methods such as bag of words, term frequency-inverse document frequency, and GloVe. 
    These are considered as three separate classification tasks and they employed traditional ML models and recurrent neural networks to predict the outcomes.
    \textcolor{black}{On the other hand, we aim to construct more specific phenotypes (e.g., `acute right frontoparietal stroke') which can be fairly easily developed from more general information extracted from the reports (e.g., extracting `hypoattenuation' in right frontoparietal distribution as well as identifying `effacement of sulci' in the same report).}

    Most of the important information, especially those describing or relating to abnormal findings, are mentioned as part of the spatial descriptions between brain imaging observations and their corresponding anatomical structures. 
    Often times, determining granular phenotypes is dependent on these specific information documented in the reports.
    Fu et al. \cite{fu2019NaturalLanguageProcessinga} developed both rule-based and ML methods to identify incidental silent brain infarct and white matter disease patients from the EHRs. 
    As reported in Fu et al.'s work, some of the false positive errors generated by the ML-based text classification system are usually contributed by certain disease locations (e.g., right occipital lobe) that often co-exist with expressions related to the disease/outcome of interest (e.g., silent brain infarct in their case).
    Thus, developing a set of constraints using domain knowledge on the spatial information in the reports has the potential to diminish such false positive cases.
    Moreover, developing constraints based on the spatial relationships between imaging observations and anatomical locations forms a natural way to predict a stroke-associated outcome of interest. 
    This also enhances the interpretability of the automatic phenotype construction system as it closely replicates a clinician's workflow to select eligible group of patients for treatment plans and clinical recommendations.
    
    Wheater et al. \cite{wheater2019ValidatedNaturalLanguagea} developed a rule-based NLP system to automatically label neuroimaging reports with a pre-defined set of 24 phenotypes. Their system incorporates manually crafted domain lexicons as well as a chunking step for extracting the radiological entities and relations from the text. Simple rules are then developed based on the presence of certain entities and relations to construct the final labels for each report. Inspired by this, we also develop the phenotype construction step similar to Wheater et al. However, our initial step of information extraction is more spatial information-oriented where we extract some common radiographic information by using advanced transformer-based language model and thus avoid the tedious process of developing manual rules for entity and relation extraction.
    
    Thus the focus of our work is to demonstrate how automatically extracted important spatial information from neuroimaging reports can potentially be used to develop granular ischemic stroke phenotypes. 
    To our knowledge, this is a first attempt toward using radiographic information connected through spatial trigger expressions in radiology reports for stroke phenotyping.
 
\section*{Materials and Methods}
\begin{SCfigure}
\includegraphics[width=0.65\textwidth]{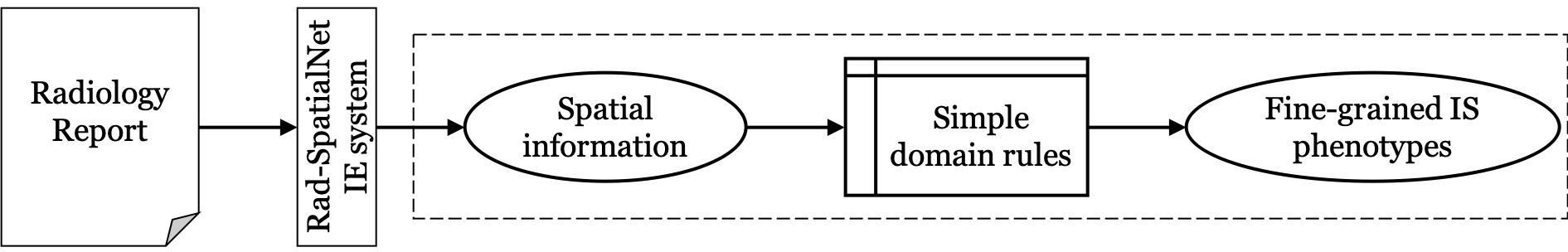}
% \vspace{-0.1in}
\centering
\caption{Pipeline for ischemic stroke (IS) phenotype classification. \textcolor{black}{Dashed box indicates the main contribution of this work. IE - information extraction.}}
\label{fig:pipeline}
\end{SCfigure}

\textcolor{black}{We use the output of a spatial information extraction (IE) system (information represented following the Rad-SpatialNet schema) to classify the granular ischemic stroke phenotypes. 
A set of simple domain rules are applied on the output of the IE system for classifying the phenotypes. 
The following sections describe the dataset along with a brief overview of the Rad-SpatialNet schema used in this study. This is followed by descriptions of the phenotype annotation process and our proposed pipeline for ischemic stroke phenotyping.
An overview of our approach is shown in Figure \ref{fig:pipeline}.}

\setcounter{section}{0}
\section{Dataset}
% \textcolor{blue}{
We select a set of \textcolor{black}{$150$} MIMIC reports (containing a mix of brain MRIs and head CTs) to classify the ischemic stroke phenotypes.
These \textcolor{black}{$150$} reports contain at least one of the ICD-9 ischemic stroke-relaed diagnosis codes from 433.01, 433.11, 433.21, 433.31, 433.81, 433.91, 434.01, 434.11, 434.91, and 436.
We refer to this phenotyping dataset as \textsc{Rad-IS-P}.
To train our spatial information extraction (IE) model, we use $400$ MIMIC-III \cite{Johnson2016mimicIII} radiology reports (consisting of chest X-rays, brain MRIs, and babygrams) annotated following Rad-SpatialNet schema as part of our earlier work \cite{datta2020RadSpatialNetFramebasedResource}.
Since we extract stroke phenotypes from both types of neuroradiology reports, i.e. MRIs and CTs, we annotated a few (15) head CT reports following the same schema to add to the training data for our spatial IE system.
Thus, we use the combined set of $415$ reports for training the IE model.
We refer to this dataset as \textsc{Rad-Spatial-IE}.

\section{Rad-SpatialNet schema}
This spatial representation schema has been proposed in our previous work \cite{datta2020RadSpatialNetFramebasedResource}.
We use this schema to extract spatial information from the \textsc{Rad-IS-P} data.
According to this, a spatial frame is constructed for each spatial description mentioned in a radiology report sentence. The spatial trigger forms the lexical unit of a spatial frame and the other related important clinical contextual information constitutes the frame elements. 
So for the sentence--`\textit{Hypodensity is noted in the pons which likely represents a lacunar infarct}', a spatial frame is instantiated by the spatial trigger `\textit{in}' and the elements associated to this lexical unit are Figure (`\textit{hypodensity}'), Ground (`\textit{pons}'), Hedge (`\textit{likely represents}'), and Diagnosis (`\textit{lacunar infarct}').
The frame elements that are not present in this example sentence but are part of the Rad-SpatialNet schema include Relative Position, Distance, Position Status, Reason, and Associated Process.
\textcolor{black}{We apply domain rules on Figure, Ground, and Diagnosis elements extracted from the reports to classify stroke phenotypes.}

\section{Ischemic stoke phenotype annotation}
\label{phenotype_annotation}
Each MRI and CT report is annotated with important IS features as validated by a practicing radiologist.
These features are identified based on both their clinical importance as well as taking into account the types of information covered in Rad-SpatialNet schema.
The pre-defined features are described as follows:
\vspace{-0.1in}
\begin{enumerate}
    \item Brain side - the laterality of the brain that is affected
    \vspace{-0.1in}
    \item \textcolor{black}{Brain region - refers to the specific brain area affected due to reduced blood and oxygen supply}
    \vspace{-0.1in}
    \item Stroke stage - three main stages used to describe the CT manifestations of stroke: acute, subacute, and chronic (as described in Birenbaum et al. \cite{birenbaum2011ImagingAcuteStroke}). \textcolor{black}{Additionally, some reports document the stage information as acute/subacute, so we also consider acute/subacute separately}
    \vspace{-0.1in}
    \item \textcolor{black}{Lacunarity - whether infarct is lacunar or not. Lacunar infarcts are usually small noncortical infarcts (diameter of 0.2 to 15 mm) and are caused by occlusion of a small perforating artery}
\end{enumerate}

Multiple combinations of these four features can be present in a report.
In such cases, we label each report with a maximum of five combinations of brain side, region, stroke stage, and lacunarity.
\textcolor{black}{For the example in Figure \ref{fig:region_specific_info}, the resulting feature combinations used for annotating the report are -- 1. right, cerebellum, acute, not lacunar, 2. left, cerebellum, chronic, not lacunar, and 3. right, brainstem (midbrain), acute, not lacunar.}
Another point to note is that if the stroke stage is directly available as part of the spatial information extracted from the report, we use that information to annotate the report, otherwise the stage annotation is determined based on certain additional conditions/domain constraints applied over the extracted spatial information.
For example, in the sentence ``\textit{There are several small acute infarctions in the right midbrain}'' in Figure \ref{fig:region_specific_info}, \textit{acute} was directly available as part of the Figure frame element \textit{acute infarctions} identified in context to the spatial trigger \textit{in}.
\textcolor{black}{However, in the last sentence, the stage is annotated as \textit{chronic} because of the presence of terms like \textit{encephalomalacia} and \textit{gliosis}.}
\textcolor{black}{Using this annotation scheme, the \textsc{Rad-IS-P} dataset was annotated with the stroke phenotypes by a radiologist (SK).}
\textcolor{black}{A brief statistics of the brain region-wise phenotype annotations are shown in Table \ref{table:region_stats}.}

\begin{SCtable}
\footnotesize
\caption{Annotated phenotypes per brain region.}
\vspace{-0.1in}
% \begin{center}
\centering
\begin{tabular}{lc|lc}
      \hline
      \textbf{Brain region affected}&\textbf{Frequency} & \textbf{Brain region affected}&\textbf{Frequency}\\
      \hline
      Cerebral hemisphere & \textcolor{black}{$26$} & Basal ganglia & \textcolor{black}{$38$}\\
      \hline
      Cerebral hemisphere - Frontal lobe & \textcolor{black}{$61$} & Thalamus & \textcolor{black}{$6$} \\
      \hline
      Cerebral hemisphere - Occipital lobe & \textcolor{black}{$30$} & Cerebral peduncle & \textcolor{black}{$2$} \\
      \hline
      Cerebral hemisphere - Parietal lobe & \textcolor{black}{$46$} & Internal/External capsule & \textcolor{black}{$8$} \\
      \hline
      Cerebral hemisphere - Temporal lobe & \textcolor{black}{$29$} & Corona radiata & \textcolor{black}{$4$} \\
      \hline
      Cerebellum & \textcolor{black}{$35$} & Insula & \textcolor{black}{$15$} \\
      \hline
      Brainstem & \textcolor{black}{$9$} & Watershed & \textcolor{black}{$4$} \\
      \hline
\end{tabular}
\label{table:region_stats}
\end{SCtable}

\section{Proposed Pipeline}
We describe the sequential stages of our phenotype extraction system in the following sections.

\vspace{-0.15in}

\subsection{Spatial information extraction}
\begin{SCfigure}
\includegraphics[width=0.45\textwidth]{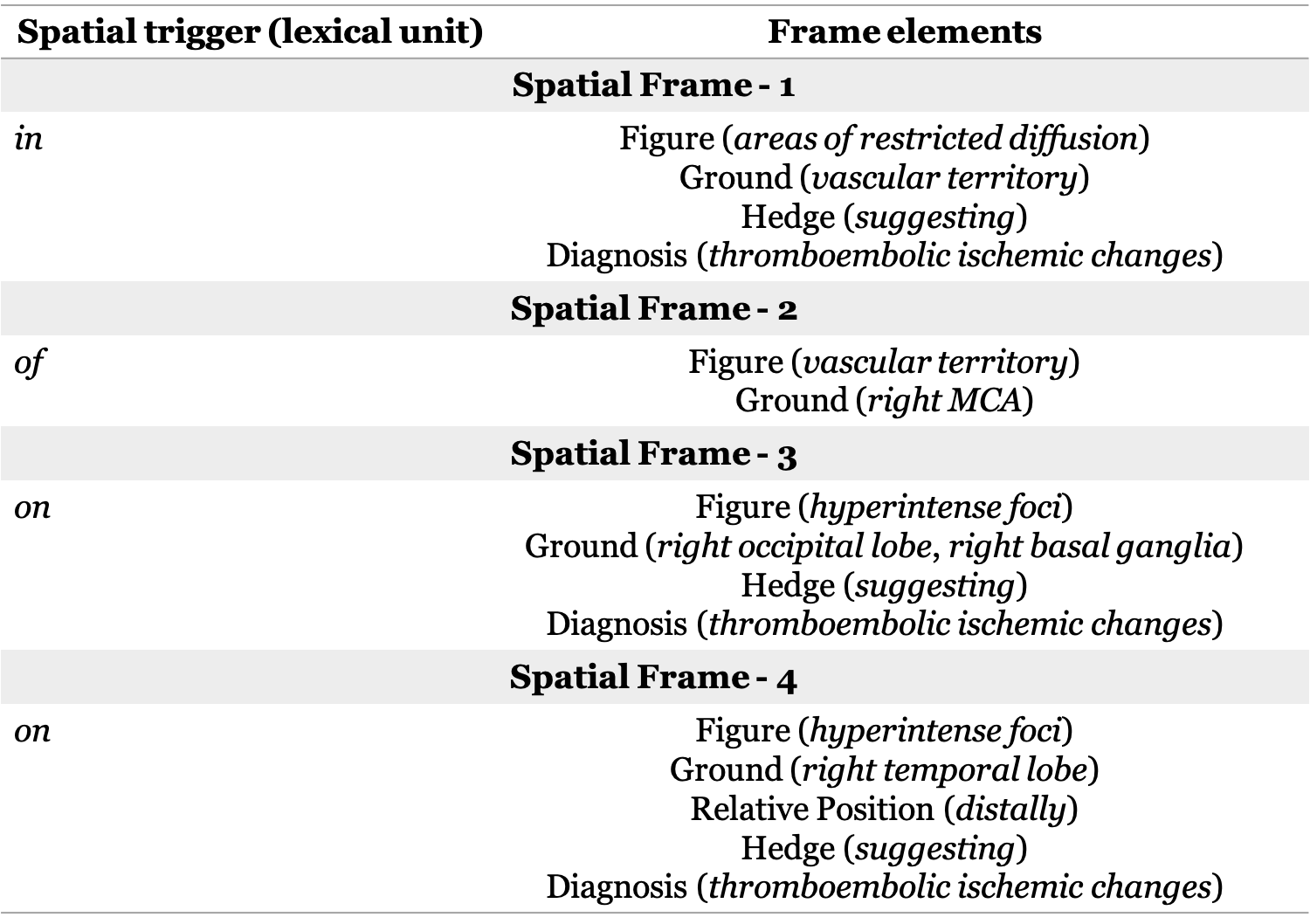}
% \vspace{-0.1in}
\centering
\caption{Spatial frames extracted for a sample sentence--\textit{There are areas of restricted diffusion \textbf{in} the vascular territory \textbf{of} the right MCA, also some scattered hyperintense foci noted \textbf{on} the right occipital lobe, right basal ganglia and distally \textbf{on} the right temporal lobe suggesting thromboembolic ischemic changes.}}
\label{fig:spatial_frames}
\end{SCfigure}

\textcolor{black}{We use an existing BERT-based sequence labeling system for extracting the spatial information from the reports \cite{datta2020RadSpatialNetFramebasedResource}. 
This includes identifying the spatial triggers in a sentence followed by identifying the associated frame elements for each extracted trigger.
The frame elements identified by the BERT system for each of the spatial triggers in a sample head CT report sentence are illustrated in Figure \ref{fig:spatial_frames}.
Specifically, in this work, we re-train the BERT-based frame element extractor using the \textsc{Rad-Spatial-IE} data with updated annotation spans for a few frame elements as described below.
}

\vspace{-0.4in}
\paragraph{\textcolor{black}{\textbf{\textit{Updates to Rad-SpatialNet for Ground and Diagnosis frame elements}}}}

Note that for each anatomical location phrase labeled as Ground element in our previous work \cite{datta2020RadSpatialNetFramebasedResource}, the associated laterality terms such as `left', `right', and `bilateral' were annotated as elements in context to that anatomical radiological entity.
Similarly, for some of the potential diagnoses labeled as Diagnosis element, the associated temporal descriptors such as `acute', `evolving', and `chronic' were also annotated as elements in context to the diagnosis radiological entity.
Thus, the laterality and the temporal descriptor terms were not part of the Ground and Diagnosis frame elements respectively (in turn not directly connected to the spatial triggers) and thus were not identified by the spatial frame element extraction system.
However, considering the need of capturing laterality and diagnosis temporality information for our phenotyping task, we updated the mention spans of the Ground and Diagnosis elements in the report sentences \textcolor{black}{to support this work}.
Consider the following examples:
\vspace{-0.1in}
\begin{enumerate}
\item Include the laterality of the anatomical location
\vspace{-0.1in}
\begin{itemize}
    \item [] Rad-SpatialNet \cite{datta2020RadSpatialNetFramebasedResource} -- \textit{There is hypodensity in the left \underline{basal ganglia}.}
    \vspace{-0.05in}
    \item [] This paper -- \textit{There is hypodensity in the \underline{left basal ganglia}.}
\end{itemize}
\vspace{-0.15in}
\item Include laterality and location descriptor whose span falls in between a laterality phrase and the anatomy phrase
\vspace{-0.25in}
\begin{itemize}
    \item [] Rad-SpatialNet \cite{datta2020RadSpatialNetFramebasedResource} -- \textit{A small area of white matter hyperintensity in the right frontal \underline{subcortical region}.}
    % could be related to prior infarct.}
    \vspace{-0.05in}
    \item [] This paper -- \textit{A small area of white matter hyperintensity in the \underline{right frontal subcortical region.}} 
    % could be related to prior infarct.}
\end{itemize}
\vspace{-0.15in}
\item Include the temporality of the potential diagnosis
\vspace{-0.1in}
\begin{itemize}
    \item [] Rad-SpatialNet \cite{datta2020RadSpatialNetFramebasedResource} --
    \textit{Hypoattenuation in the right frontoparietal distribution consistent with acute \underline{infarction}.}
    \vspace{-0.2in}
    \item [] This paper -- \textit{Hypoattenuation in the right frontoparietal distribution consistent with \underline{acute infarction}.}
\end{itemize}
\end{enumerate}
\vspace{-0.1in}
In the first example we see that `\textit{left}' has been included in the Ground element, and in the second example both `\textit{right}' and `\textit{frontal}' are included in the Ground element span.
In the third example,`\textit{acute}' is included in the Diagnosis element span.
The spatial trigger (lexical unit for a spatial frame) is `\textit{in}' for all the examples.  

\vspace{-0.1in}
\subsection{Automatic IS phenotype extraction}
For each report, we use rules on top of the output of the BERT-based element extractor to automatically classify the phenotypes.
We combine the spatial frames identified by the element extractor at the report level.
We also keep a track of all the spatial frames predicted by the BERT extractor for each sentence in a sequential order (the order in which the spatial triggers appear in a sentence).
\textcolor{black}{This helps to combine the frames when the Ground element associated to a trigger is same as the Figure element of the next trigger.
For example, in ``\textit{acute infarction in the lateral aspect of right cerebellum}'', IS-related finding (\textit{infarction}) is connected to the corresponding location (\textit{right cerebellum}) through the common frame element \textit{aspect} of the two spatial frames with triggers \textit{in} and \textit{of} appearing sequentially in the sentence.
}
For each spatial trigger identified in a sentence, the following steps are performed:
\vspace{-0.1in}
\begin{enumerate}
    \item \textcolor{black}{First, the spatial triggers and the frame elements relevant to ischemic stroke are filtered. For this, we check if any of the Figure/Diagnosis element spans detected in relation to a trigger is IS-related. If one of the pre-defined IS-related imaging finding keywords (as shown in the first two rows of Table \ref{table:keywords}) is present in any of the element spans, the following steps are performed.}
    \vspace{-0.1in}
    \item For extracting the brain side, we check for the presence of any laterality-related term in the predicted Ground element span \textcolor{black}{(e.g., \textit{left} for left, and \textit{both}, \textit{bilateral} for bilateral). Additionally, if the Ground elements are \textit{thalami} and \textit{capsules}, we assign the side as \textit{bilateral}. In other cases, \textit{unspecified} is assigned.
    Moreover, in cases (e.g., \textit{infarction involving left frontal and parietal lobes}) when the same laterality is linked to multiple regions, each region is assigned the laterality separately. Here, \textit{left} is assigned to both \textit{frontal} and \textit{parietal} lobes although \textit{left} does not appear in the Ground span \textit{parietal lobes}.}
    \vspace{-0.1in}
    \textcolor{black}{\item For identifying the brain region, the presence of any of the keywords developed for each of the pre-defined brain areas are checked in the detected Ground element span
    (e.g., keywords for mapping the brain region as \textbf{Basal ganglia} are--\textit{basal ganglia}, \textit{caudate}, \textit{caudate nucleus}, \textit{caudate head}, \textit{caudate nucleus head}, \textit{putamen}, \textit{globus pallidus}, and \textit{lentiform nucleus}). These keywords are built with domain expert input. Additionally, for Ground element spans involving two lobes, we assign both the cerebral lobes (e.g., \textit{frontal} and \textit{parietal} lobes are assigned for Ground element span--\textit{frontoparietal}).}
    \vspace{-0.1in}
    \item For identifying the stroke stage \textcolor{black}{for each pair of brain region and side}, two sequential steps are involved. First, we check for the presence of any stage-related term directly in the predicted Figure/Diagnosis element span.
    \textcolor{black}{Since the term \textit{acute} is also contained in \textit{subacute}, we priortize the search for subacute over acute.}
    If not found, domain constraints are applied over the predicted spatial frame elements (this step also takes into account the other spatial relationships predicted in the same report in connection to the same brain region).
    \textcolor{black}{If the stage is not determined by these two steps, we assign the label-- \textit{Can't determine}.}
    \vspace{-0.1in}
    \item \textcolor{black}{Similarly, for identifying the lacunarity for each pair of brain region and side, we check for the presence of lacunar-specific terms in the Figure/Diagnosis element span.
    We assign a binary lacunarity label--\textit{Yes} if lacunar and \textit{No} otherwise.}
\end{enumerate}

\textcolor{black}{The keywords developed for IS-related imaging findings as well as for identifying the stroke stage and lacunarity from the frame element spans are shown in Table \ref{table:keywords}.}
\textcolor{black}{These keywords as well as the domain constraints for inferring the stage are developed in collaboration with the radiologist who created the gold phenotypes.}
A few predominant constraints are demonstrated in Table \ref{table:domain_constraints}.

\begin{table}[t]
\footnotesize
\caption{Keywords for identifying IS finding, IS stage, and lacunarity from the frame element spans to classify the phenotypes.}
\vspace{-0.25in}
\begin{center}
\resizebox{\textwidth}{!}{
\begin{tabular}{l|p{0.75\columnwidth}}
      \hline
      \textbf{Item}&\textbf{Keywords} \\
      \hline
      IS-related imaging finding (CT) & hypodensity, hypodensities, hyperdensity, hyperdensities, hypodense, hypoattenuation, hypo-attenuation, low attenuation, low-attenuation, hypoattenuating, hypo-attenuating, low attenuating, low-attenuating, decreased attenuation, lacune, infarct, lesion \\
      \hline
      IS-related imaging finding (MRI) & restricted diffusion, slow diffusion, susceptibility artifact, signal, infarct \\
      \hline
      IS stage - Subacute & sub-acute, subacute, sub acute, evolving \\
      \hline
      IS stage - Acute & acute \\
      \hline
      IS stage - Chronic & encephalomalacia, gliosis, known, old, previous, prior \\ 
      \hline
      Lacunarity & lacune, lacunar \\
      \hline
\end{tabular}}
\end{center}
\label{table:keywords}
\end{table}

\begin{table}[t]
% \footnotesize
\caption{Domain constraints applied on BERT predicted spatial frame elements to determine ischemic stroke stage.}
\vspace{-0.25in}
\begin{center}
\resizebox{\textwidth}{!}{
\begin{tabular}{l|c|c}
      \hline
      \textbf{Modality}&\textbf{Acute}&\textbf{Chronic} \\
      \hline
      CT & \Longstack{(hypodensity/hypoattenuation in cortical/subcortical region) \\ AND (hyperdense MCA OR hyperdensity in basilar artery OR \\ loss of gray-white matter differentiation OR sulcal effacement)} & \Longstack{(hypodensity/hypoattenuation in cortical/subcortical region \\ AND (prominence of ventricles/sulci OR atrophy)) OR gliosis/encephalomalacia} \\
      \hline
      MRI & \Longstack{(slow diffusion/restricted diffusion in cortical/subcortical region) \\ OR (loss of flow void in MCA/basilar artery)} & \Longstack{facilitated diffusion in cortical/subcortical region \\ OR gliosis/encephalomalacia OR dilation of ventricles}\\
      \hline
\end{tabular}}
\end{center}
\label{table:domain_constraints}
\end{table}

\section*{Experimental Settings and Evaluation}
We use the BERT\textsubscript{LARGE} model for fine-tuning the spatial information extraction task by initializing the model parameters obtained after pre-training BERT on MIMIC-III clinical notes for $ 300,000 $ steps \cite{si2019EnhancingClinicalConcept}.
For extracting the spatial triggers from the \textsc{Rad-IS-P} data, we use the trained model from our previous work \cite{datta2020RadSpatialNetFramebasedResource}.
However, for extracting the frame elements, we re-train the BERT-based element extractor on the \textsc{Rad-Spatial-IE} dataset using the updated gold spans of Ground and Diagnosis frame elements for capturing the laterality and temporality information, respectively.
We split the reports in \textsc{Rad-Spatial-IE} into training, validation, and test sets in the ratio of 80-10-10\% and perform 10-fold cross-validation for evaluating the performance of the element extractor model.
The model is fine-tuned by setting the maximum sequence length at $128$, learning rate at $2e-5$, and number of training epochs at $4$. We use cased version of the models.
Among the $10$ versions of the trained model checkpoints (generated for 10 folds of the dataset), we select the version based on the highest F1 measure on the validation set to predict the spatial frame elements from the \textsc{Rad-IS-P} data used for phenotype classification.
Additionally, to provide a sense of the performance of the spatial information extraction system on stroke-related reports (that are more representative of the ones used for phenotyping), we annotated a random set of $20$ reports from the \textsc{Rad-IS-P} dataset according to the Rad-SpatialNet schema and evaluated the system's performance on these $20$ reports.
\textcolor{black}{For our phenotyping task, we report the precision, recall, and F1 measures of the automatic phenotype extraction system based on various meaningful subsets or combinations of stroke features described in Section \ref{phenotype_annotation}.}

\section*{Results}
The average precision, recall, and F1 scores of extracting spatial triggers from the \textsc{Rad-Spatial-IE} data are $86.14$\%, $79.55$\%, and $82.66$, respectively.
For the $20$ stroke reports (selected from the \textsc{Rad-IS-P} data), the precision, recall, and F1 values for spatial trigger extraction are $93.70$\%, $76.28$\%, and $84.10$, respectively.
These predicted triggers are used further by the element extractor model in the end-to-end evaluation (shown under the `Predicted spatial triggers' column in Table \ref{tab:results_rel_large}).
Table \ref{tab:results_rel_large} also highlights the average 10-fold CV performance measures of the BERT-based element extractor using the gold spatial triggers.
\textcolor{black}{The frame elements Associated Process and Reason have very low performance scores as they occur very rarely in the whole dataset and also not used for phenotyping.}
We additionally illustrate the overall precision, recall, and F1 measures (considering all the spatial frame elements) of the frame element extractor on the $20$ stroke report subset in Table \ref{tab:sprels_results_20_reports}.

\textcolor{black}{The results of our phenotype extraction system are shown in Table \ref{tab:phenotyping_results}.
We calculate the performance metrics of the system based on different combinations of the features (i.e., brain region, side, stroke stage, and lacunarity) that are potentially useful for clinical research studies.
The precision, recall, and F1 values are calculated by comparing the distinct combinations of the features per report identified by the system to those of the gold annotated ones.
This gives an idea about how well the system performs in classifying various subsets of meaningful features.
Since stroke stage and lacunarity are associated with a specific brain region and side pair, we report the performance of the system including the stage and lacunarity features along with brain region and side in the last four rows of the table.
Note that for stroke stage, we show the results both by considering various stage types and also by grouping the three stage types--acute, subacute, and acute/subacute together.}

\begin{table*}[t]
\footnotesize
\caption{10 fold CV results on \textsc{Rad-Spatial-IE} for BERT-based spatial frame element extraction model using gold and predicted spatial triggers. P - Precision, R - Recall.}
\vspace{-0.1in}
\centering
	\begin{tabular}{lccc|ccc}
		\hline 
		\multirow{2}{*}{\textbf{Main Frame Elements}} &
		\multicolumn{3}{c}{\textbf{Gold spatial triggers}} &
		\multicolumn{3}{c}{\textbf{Predicted spatial triggers}} \\
		\cline{2-7}
		& \textbf{P(\%)} & \textbf{R(\%)} & \textbf{F1} & \textbf{P(\%)} & \textbf{R(\%)} & \textbf{F1} \\
		\hline 
		\textsc{Figure} & 81.39 &  84.26 & 82.77 &  67.53 & 71.08 & 69.14 \\
		\hline
		\textsc{Ground} & 92.01 &  93.41 &  92.69  & 70.87 & 80.13 & 75.09 \\
		\hline
		\textsc{Hedge} & 75.51 & 83.08 & 78.91 & 68.94 & 74.05 & 71.19 \\
		\hline
		\textsc{Diagnosis} &  54.73 & 78.41 & 64.06 & 48.49 & 67.67 &  55.95 \\
		\hline
		\textsc{Relative Position} & 87.47 & 81.01 & 83.54 & 60.13 & 66.35 & 62.17 \\
		\hline
		\textsc{Distance} & 75.83 & 80.83 & 75.53 & 73.63 &  80.00 & 74.25 \\
		\hline
		\textsc{Position Status} & 68.59 & 66.20 &  66.97 & 61.42 & 64.45 & 61.55 \\
		\hline
% 		\textsc{Associated Process}, \textsc{Reason} & 0.0 & 0.0 & 0.0 & 0.0 & 0.0 & 0.0 \\
% 		\hline
		\textsc{Overall} & 82.60 & 85.31 & 83.92 & 66.95 & 73.17 & 69.81 \\
		\hline
	\end{tabular}
\label{tab:results_rel_large}
\vspace{0.1in}
\end{table*}

\begin{table*}[t]
\footnotesize
% \vspace{-0.1in}
\caption{BERT-based spatial frame element extractor's performance on 20 stroke reports (taken from \textsc{Rad-IS-P}). P - Precision, R - Recall.}
\vspace{-0.15in}
\centering
\begin{tabular}{lccc}
      \hline
      \textbf{Spatial triggers used}&\textbf{Overall P (\%)}&\textbf{Overall R (\%)}&\textbf{Overall F1} \\
      \hline
      Gold annotated triggers & 72.80 & 80.87 & 76.62  \\
      \hline
      Predicted triggers & 65.71 & 73.48 & 69.38 \\
      \hline
\end{tabular}
\label{tab:sprels_results_20_reports}
\vspace{0.1in}
\end{table*}

\begin{table*}[t]
\footnotesize
\caption{Phenotype extraction results. BR - brain region, CS - corresponding side, SS - stroke stage, SS\_CO - SS with coarse types (\textit{acute}/\textit{chronic}), LC - lacunarity.}
\vspace{-0.1in}
\centering
% \resizebox{\textwidth}{!}{
\begin{tabular}{l|l|c|c|c}
      \hline
      \textbf{Phenotype variant} & \textbf{Example} & \textbf{Precision(\%)} & \textbf{Recall(\%)} & \textbf{F1} \\
      \hline
      BR & \textit{cerebellum} & \textcolor{black}{$73.58$} & \textcolor{black}{$89.62$}  & \textcolor{black}{$80.81$} \\
      \hline
      BR $+$ CS & \textit{cerebellum, left} & \textcolor{black}{$68.34$} & \textcolor{black}{$85.47$} & \textcolor{black}{$75.95$} \\
      \hline
      BR $+$ SS\_CO & \textit{cerebellum, chronic} & \textcolor{black}{$55.53$} & \textcolor{black}{$82.0$} & \textcolor{black}{$66.22$} \\
      \hline
      BR $+$ CS $+$ SS\_CO & \textit{cerebellum, left, chronic} & \textcolor{black}{$49.67$} & \textcolor{black}{$74.11$} & \textcolor{black}{$59.48$} \\
      \hline
      BR $+$ CS $+$ SS & \textit{cerebral hemisphere - frontal lobe, bilateral, subacute} & \textcolor{black}{$46.32$} & \textcolor{black}{$56.96$} & \textcolor{black}{$51.09$} \\
      \hline
      BR $+$ CS $+$ LC & \textit{basal ganglia, bilateral, yes} & \textcolor{black}{$62.53$} & \textcolor{black}{$77.2$} & \textcolor{black}{$69.09$} \\
      \hline
      BR $+$ CS $+$ SS\_CO $+$ LC & \textit{basal ganglia, bilateral, chronic, yes} & \textcolor{black}{$48.59$} & \textcolor{black}{$72.49$} & \textcolor{black}{$58.18$} \\
      \hline
\end{tabular}
% }
\label{tab:phenotyping_results}
\end{table*}

\section*{Discussion}

\textcolor{black}{
This work focuses on identifying complex ischemic stroke phenotypes mainly from the perspective of the stroke location (brain region and side).
We utilize the output of a spatial information extraction (IE) system (developed in our previous work) and apply simple neuroradiology-specific rules to classify these phenotypes.
Note that the phenotypes we tackle in this work consider information at the level of specific brain area that is affected by stroke.
Thus, this involves identification of information related to a stroke affected region in the brain from the report text.
Our Rad-SpatialNet schema allows for easy identification of such related information as this captures the spatial relations between imaging findings and brain locations as well as the associated potential diagnoses.
This becomes even more useful when the same report contains infarcts of different stages in different brain locations.
Figure \ref{fig:region_specific_info} illustrates an example where three different brain regions are affected and the stroke stage varies according to the region and its laterality.
}

\newpage
\textcolor{black}{
We observe that applying simple domain rules that are mainly based on keyword search and a small set of constraints over the output of the spatial IE system results in satisfactory performance in classifying complex stroke phenotypes.
This highlights both the information coverage of the Rad-SpatialNet schema and the sufficiently promising performance of the spatial IE system. Another point to note is that the information covered through Rad-SpatialNet are generic enough to extend our phenotype classification approach to other types of diseases/conditions beyond neuroradiology domains.
}

\textcolor{black}{
We briefly discuss the errors of the phenotype extraction system here.
Most of the errors related to missing the brain region (referring to the recall of \textcolor{black}{$89.62$\%} in Table \ref{tab:phenotyping_results}) is because of the Ground elements that are not predicted by the spatial IE system.
There are also a very few cases where spatial triggers are not present explicitly (e.g., \textit{left cerebellar infarct}). The existing Rad-SpatialNet schema doesnot capture such implicit relations and thus such regions are missed.
Some of the errors related to stroke stage classification (when all the stage types are considered) is due to the ambiguity involved in distinguishing the acute and the subacute stages. Oftentimes, it becomes difficult to assess the stroke timing based on the report content}
\textcolor{black}{(one of the major reasons for low recall for BR $+$ CS $+$ SS shown in Table \ref{tab:phenotyping_results}).}
\textcolor{black}{
A small number of errors also occur when only acute and chronic stage information is considered because the output of the spatial IE system sometimes missed the specific stage-related term (e.g., \textit{evolving}, \textit{chronic}) in the predicted Diagnosis/Figure element span.
Moreover, the report does not contain other spatial relations to satisfy the domain constraints for stage inference.
Another reason of stage-related errors is when the stage information is mentioned in a following sentence in the report that does not contain any spatial relations (e.g., `\textit{These lesions suggest old infarction'}).
Lacunar-related errors happen mainly because their inferences sometimes depend on the specific sizes mentioned in the sentence (e.g., \textit{lesion of 7 mm in diameter}) that are currently not captured in the Rad-SpatialNet schema.
Taking into account a few limitations as described here in the Rad-SpatialNet schema, we aim to emphasize that there are rare instances of such scenarios overall across reports and we intend to further incorporate these information in the Rad-SpatialNet in our future work.
\textcolor{black}{We also see that the precision values are low, and one of the main reasons is that many of the stroke locations are referenced multiple times in a report and are expressed differently or with varying levels of specificity. For example, \textit{left frontal lobe} is mentioned in the report's Findings section, whereas \textit{left MCA} is mentioned in the Impressions section. This results in generating some false positive brain regions (e.g., \textit{parietal} and \textit{insula} here) as MCA (middle cerebral artery) maps to parts of \textit{frontal} and \textit{parietal lobes} as well as \textit{insula} (the brain regions where MCA supplies blood to).}
The performance of our phenotype extraction system reflects the challenging nature of this complex phenotyping task and we aim to improve its performance and evaluate on an augmented dataset in a later work.
}

\textcolor{black}{
However, the phenotyping results suggest that the Rad-SpatialNet schema that we used in this work is robust enough considering the complexity of the phenotypes.
We want to highlight that the current Rad-SpatialNet schema can be leveraged further to classify more granular aspects of the stroke location. Specifically, the RelativePosition frame element (e.g., \textit{superior}, \textit{inferior}) can be used to classify the subregions of a brain region like \textit{cerebellum}. For instance, in the sentences of the same report--``\textit{New acute infarction involving the superior left cerebellar hemisphere}'' and ``\textit{Encephalomalacia and gliosis are again seen in the inferior left cerebellar hemisphere}'', the stroke stage is \textit{acute} in case of left cerebellum (superior) and \textit{chronic} for left cerebellum (inferior).
Thus, spatial information documented in the reports when extracted with detailed contextual information facilitates the classification of fine-grained phenotypes.
}

\section*{Conclusion}

\textcolor{black}{
We used the output of an existing spatial information extraction system based on the Rad-SpatialNet schema to classify complex IS phenotypes.
We demonstrated that a generalizable and fine-grained representation schema like Rad-SpatialNet could be utilized for determining detailed phenotypes that often requires information about various related radiological entities (such as findings, brain locations, and diagnoses).
Our phenotypes are mainly based on specific brain regions affected by stroke.
We have shown that satisfactorily good results
% (recall of \textcolor{red}{$80.6$\%}) 
can be achieved by applying simple domain rules on top of the IE system's output to classify the phenotypes.
}

\vspace{-0.2in}
\paragraph{Acknowledgments}

\textcolor{black}{This work was supported in part by the National Institute of Biomedical Imaging and Bioengineering (NIBIB: R21EB029575) and the Patient-Centered Outcomes Research Institute (PCORI: ME-2018C1-10963).}

% \end{thebibliography}

% \bibliographystyle{vancouver}
\bibliographystyle{vancouver-mod}
% \vspace{-0.1in}
\setlength{\bibsep}{0.2pt}

\bibliography{summit2021}

\end{document}